\documentclass[12pt]{article}

\usepackage{booktabs}
\usepackage{siunitx}
\newcommand{\mc}[2]{#1\,{$_{\pm#2}$}} 
\newcommand{\best}[1]{\textbf{#1}}        
\newcommand{\second}[1]{\underline{#1}}   

\newcommand{\hr}{\mathbf{x^{HR}}}
\newcommand{\hathr}{\mathbf{\hat{x}^{HR}}}
\newcommand{\lr}{\mathbf{x^{LR}}}

\usepackage{etoolbox}
\usepackage{bm}

\usepackage{arydshln}

\makeatletter
\def\mathcolor#1#{\@mathcolor{#1}}
\def\@mathcolor#1#2#3{%
	\protect\leavevmode
	\begingroup
	\color#1{#2}#3%
	\endgroup
}
\makeatother

\usepackage{url}

\usepackage[table,xcdraw]{xcolor}

\usepackage{graphicx}
\usepackage{multirow}

\graphicspath{ {./figures/} }

\usepackage{caption}
\usepackage{subcaption}

\usepackage{amsmath} 
\usepackage{amsfonts}
\usepackage{nicefrac}

\usepackage[a4paper, total={6in, 8in}]{geometry}

\usepackage[symbol]{footmisc}

\usepackage{multirow}

\usepackage[
colorlinks=true,
linkcolor=blue,
filecolor=blue,
urlcolor=blue,
citecolor=blue,
pagebackref=true
]{hyperref}
\makeatletter
\renewcommand*{\backref}[1]{}
\renewcommand*{\backrefalt}[4]{%
	\ifcase #1
	(Not cited.)%
	\or
	\textsuperscript{#2}%
	\else
	\textsuperscript{#2}%
	\fi}
\makeatother

\begin{document}
	\begin{center}	
\sf {\Large {\bfseries Efficient Vision Mamba for MRI Super-Resolution via Hybrid Selective Scanning}} \\ 
Mojtaba Safari$ ^1 $, Shansong Wang$ ^1 $, Vanessa L Wildman$ ^1 $, Mingzhe Hu$ ^1 $, Zach Eidex$ ^1 $,  Chih-Wei Chang$ ^1 $, Erik H Middlebrooks$ ^2 $, Richard L.J Qiu$ ^1 $, Pretesh Patel$ ^1 $,  Ashesh B. Jani$ ^1 $, Hui Mao$ ^3 $, Zhen Tian$ ^4 $, and Xiaofeng Yang$ ^{1, \ddagger} $ \\
%
\end{center}
{$ ^1 $Department of Radiation Oncology and Winship Cancer Institute, Emory University, Atlanta, GA, United States.\\
$ ^2 $Department of Radiology, Mayo Clinic, Jacksonville, FL, United States of America.\\
$ ^3 $Department of Radiology and Imaging Science and Winship Cancer Institute, Emory University, Atlanta, GA, USA \\
$ ^4 $Department of Radiation and Cellular Oncology, University of Chicago, Chicago, IL, United States.\\
$ ^\ddagger  $Corresponding Author: email: \url{xiaofeng.yang@emory.edu}}\\

\newpage

\begin{abstract}
	\noindent\textbf{\textit{Background.}} High-resolution MRI is essential for accurate diagnosis and treatment planning, but its clinical acquisition is often constrained by long scanning times, which increase patient discomfort and reduce scanner throughput. While super-resolution (SR) techniques offer a post-acquisition solution to enhance resolution, existing deep learning approaches face trade-offs between reconstruction fidelity and computational efficiency, limiting their clinical applicability.
	
	\noindent\textbf{\textit{Purpose.}} This study aims to develop an efficient and accurate deep learning framework for MRI super-resolution that preserves fine anatomical detail while maintaining low computational overhead, enabling practical integration into clinical workflows.
	
	\noindent\textbf{\textit{Materials and Methods.}} We propose a novel SR framework based on multi-head selective state-space models (MHSSM) integrated with a lightweight channel multilayer perceptron (MLP). The model employs 2D patch extraction with hybrid scanning strategies (vertical, horizontal, and diagonal) to capture long-range dependencies while mitigating pixel forgetting. Each MambaFormer block combines MHSSM, depthwise convolutions, and gated channel mixing to balance local and global feature representation. The framework was trained and evaluated on two distinct datasets: 7T brain T1 MP2RAGE maps (142 subjects) and 1.5T prostate T2w MRI (334 subjects). Performance was compared against multiple baselines including Bicubic interpolation, GAN-based (CycleGAN, Pix2pix, SPSR), transformer-based (SwinIR), Mamba-based (MambaIR), and diffusion-based (I\textsuperscript{2}SB, Res-SRDiff) methods.
	
	\noindent\textbf{\textit{Results.}} The proposed model demonstrated superior performance across all evaluation metrics while maintaining exceptional computational efficiency. On the 7T brain dataset, our method achieved the highest structural similarity (SSIM: $0.951 \pm 0.021$) and peak signal-to-noise ratio (PSNR: $26.90 \pm 1.41$ dB), along with the best perceptual quality scores (LPIPS: $0.076 \pm 0.022$; GMSD: $0.083 \pm 0.017$). These results represented statistically significant improvements over all baselines ($p < 0.001$), including a 2.1\% SSIM gain over SPSR and a 2.4\% PSNR improvement over Res-SRDiff. For the prostate dataset, the model similarly outperformed competing approaches, achieving SSIM of $0.770 \pm 0.049$, PSNR of $27.15 \pm 2.19$ dB, LPIPS of $0.190 \pm 0.095$, and GMSD of $0.087 \pm 0.013$. Notably, our framework accomplished these results with only 0.9 million parameters and 57 GFLOPs, representing reductions of 99.8\% in parameters and 97.5\% in computational operations compared to Res-SRDiff, while also substantially outperforming SwinIR and MambaIR in both accuracy and efficiency metrics.
	
	\noindent\textbf{\textit{Conclusion.}} The proposed framework provides a computationally efficient yet accurate solution for MRI super-resolution, delivering well-defined anatomical details and improved perceptual fidelity across anatomically distinct datasets. By significantly reducing computational demands while maintaining state-of-the-art performance, the model offers strong potential for feasibility toward clinical translation and scalable integration into future imaging workflows.
\end{abstract}

\textbf{\textit{Keywords:}} {MRI, Deep learning, ultra high field MRI, super-resolution, state-space model, SSM}

\newpage 
\section{Introduction}

Magnetic resonance imaging (MRI) is an essential used modality in both clinical and research settings due to its excellent soft-tissue contrast, high acquisition flexibility, and lack of ionizing radiation exposure. For example, quantitative 3D magnetization-prepared 2 rapid acquisition gradient echo (MP2RAGE) T1 mapping eliminates receptive field bias and first-order transmit field inhomogeneities in brain imaging. These maps enable precise diagnostic and treatment planning by Identifying hypoxic regions that can guide adaptive dose-painting radiotherapy~\cite{park2024optimization,epel2019oxygen}. Likewise, anatomical T2-weighted (T2w) MRI provides superior soft-tissue contrast, making it indispensable for prostate cancer detection, staging, treatment planning, and longitudinal surveillance~\cite{barrett2019pi}. These examples underscore the importance of achieving high-resolution MRI across diverse anatomical sites and clinical tasks.

However, MRI resolution is fundamentally limited by hardware constraints and the extended acquisition times required for fine spatial detail. Higher resolution demands longer scans, which increase patient discomfort, heighten the risk of motion artifacts, and reduce scanner throughput~\cite{SAFARI2026105704, https://doi.org/10.1002/mp.17675}. Furthermore, high-resolution imaging typically requires high-field strength scanners (3T and above), which remain costly and inaccessible to many healthcare settings.  To address these limitations, deep learning (DL)-based reconstruction methods have achieved impressive results, but typically requiring access to raw $k$-space data, under-sampling masks, and coil sensitivity maps~\cite{SAFARI2026108291}. In practice, access to raw $k$-space data might be limited by multiple factors such as vendor restrictions, proprietary data formats, and institutional privacy regulations. In contrast, DL-based super-resolution (SR) approaches attempt to recover high-resolution (HR) images from low-resolution (LR) inputs without such priors, offering a more flexible and widely applicable solution~\cite{YU2025110345}.

DL has revolutionized MRI SR, with methodologies evolving through several distinct paradigms, each addressing core trade-offs between fidelity, perceptual quality, and computational efficiency \cite{khateri2025mri, qiu2023medical}. Early convolutional neural network (CNN)-based approaches, such as EDSR \cite{lim2017enhanced} and RCAN \cite{zhang2018image}, established strong baselines by leveraging deep residual learning to enhance spatial detail. However, their inherently limited receptive field constrains the modeling of long-range spatial dependencies, which are crucial for global anatomical consistency. Generative Adversarial Network (GAN)-based methods improved perceptual realism but often at the cost of introduced hallucinations and training instability \cite{isola2017image, zhu2017unpaired}. The advent of Vision Transformers (ViTs) addressed the context modeling limitation through self-attention mechanisms, enabling superior performance in tasks like medical image SR (e.g., TransMRSR \cite{huang2023transmrsr}). Nonetheless, the quadratic computational complexity of self-attention with respect to input size remains a significant bottleneck for high-resolution 3D medical images. Efficient variants like SwinIR adopted window-based attention to mitigate this cost \cite{liang2021swinir}. More recently, diffusion models have set new state-of-the-art benchmarks in generative SR by iteratively denoising images, though they require extensive computational resources and sampling steps \cite{Safari_2025_res_SRDiff,yue2024efficient}. The state-space model (SSM) paradigm, exemplified by Mamba, has emerged as a highly efficient alternative for long-sequence modeling, achieving linear-time complexity through selective scanning mechanisms \cite{gu2023mamba}. Initial adaptations for vision tasks, such as MambaIR, show promise but can be susceptible to pixel forgetting in standard 2D scanning patterns \cite{10.1007/978-3-031-72649-1_13}.

Building directly on the SSM paradigm, vision Mamba architectures such as MambaIR \cite{10.1007/978-3-031-72649-1_13} apply selective scanning strategies for efficient image restoration. Similar to ViTs, these methods represent images as sequences of patches but employ recurrent state updates to implicitly capture long-range dependencies with reduced memory overhead. However, existing implementations can be susceptible to pixel forgetting when forming horizontal and vertical tokens \cite{10.1007/978-3-031-72649-1_13}, and densely sampled input patches may impose additional computational burden.

In this study, we propose an efficient Vision Mamba framework for MRI SR. Our contributions are three-fold: (i) a hybrid selective scanning strategy (vertical, horizontal, diagonal) to mitigate pixel forgetting and enhance long-range dependency modeling, (ii) integration of a lightweight channel MLP to reduce parameter overhead while preserving representational power, and (iii) application and validation on two distinct datasets: 7T MP2RAGE brain T1 maps and 1.5T prostate T2w images. By combining efficiency with high-fidelity reconstruction, the proposed framework provides a solution to bridge the gap between research-oriented SR models and potential clinical utility.

\section{Materials and methods}\label{sec:material_methods}

\subsection{Vision Mamba}

\begin{figure}[t!]
	\centering
	\includegraphics[width=0.8\textwidth, draft=false]{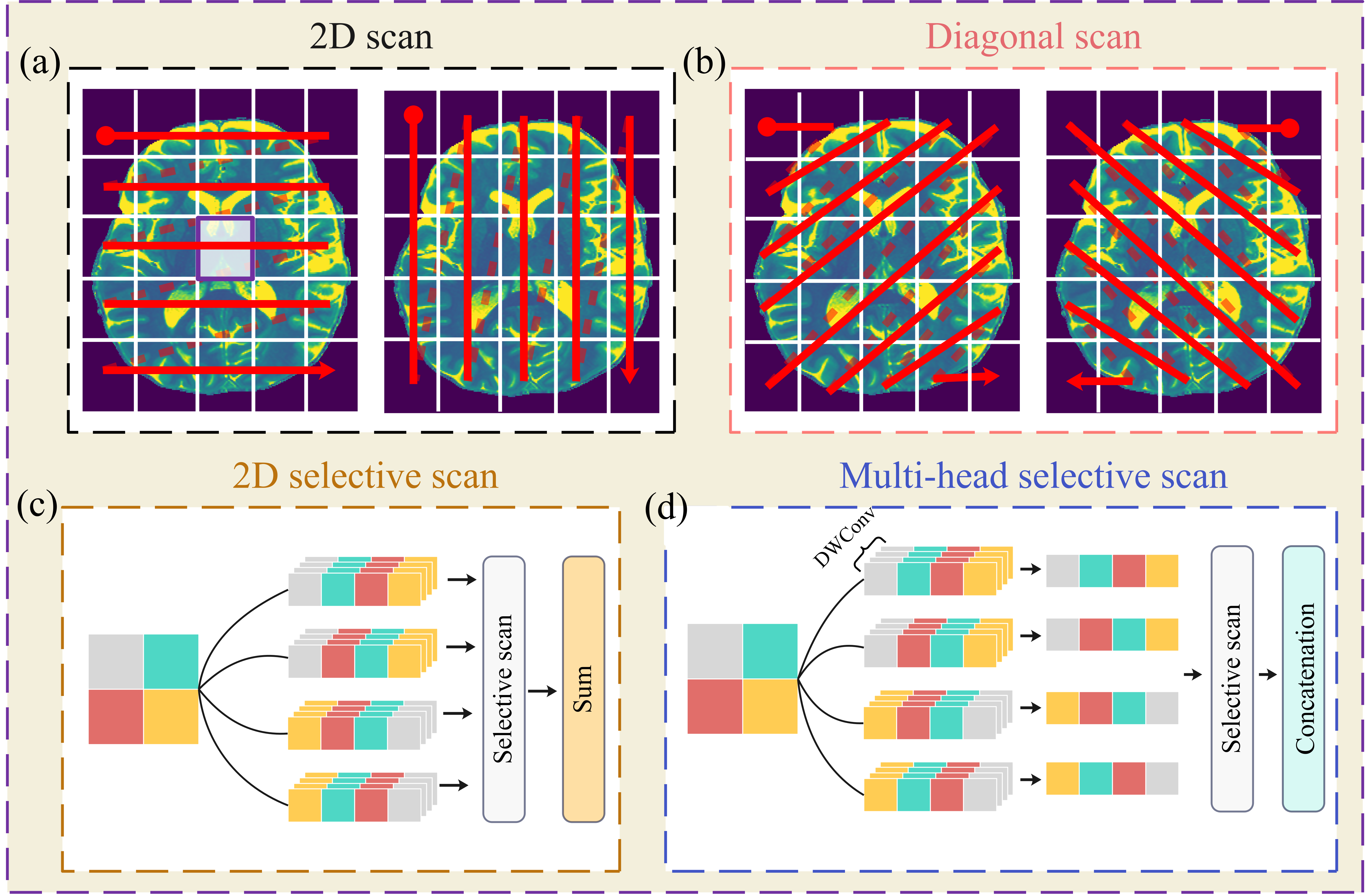}
	\caption{Selective scanning strategies in Vision Mamba. (a) Horizontal and vertical scanning may separate central pixels from their diagonal neighbors. (b) Diagonal scanning preserves spatial adjacency. (c) Dense extraction increases parameter overhead. (d) Our efficient variant uses depthwise convolutions for preprocessing.}
	\label{fig:selective_scan}
\end{figure}

Similar to the ViT, the Vision Mamba architecture partitions an input image into non-overlapping 2D patches, which are then processed through a SSM using selective scanning strategies (Figures~\ref{fig:selective_scan}(a)-(b)). Conventional approaches adopt horizontal and vertical scans~\cite{liu2024vmamba,zhu2024vision}, but these separate central pixels (white box in Figure~\ref{fig:selective_scan}(a)) from their diagonal neighbors, thereby limiting the ability of Vision Mamba models to capture long-range spatial dependencies. In contrast, diagonal scanning preserves adjacency between central and diagonal pixels (Figure~\ref{fig:selective_scan}(b))~\cite{lin2025eamamba,xu2024visual}. In this study, we employ a hybrid strategy that combines vertical, horizontal, and diagonal scans to form the input sequences.

Densely extracted patches require separate sets of parameters, which can increase computational cost of selective scanning (Figure~\ref{fig:selective_scan}(c)). To address this, our efficient variant first applies depthwise convolutional layers to preprocess patches, reducing the number of trainable parameters while maintaining representational capacity. The outputs are concatenated to form the input sequence to the SSM (Figure~\ref{fig:selective_scan}(d)).

\begin{figure}[t!]
	\centering
	\includegraphics[width=\textwidth, draft=false]{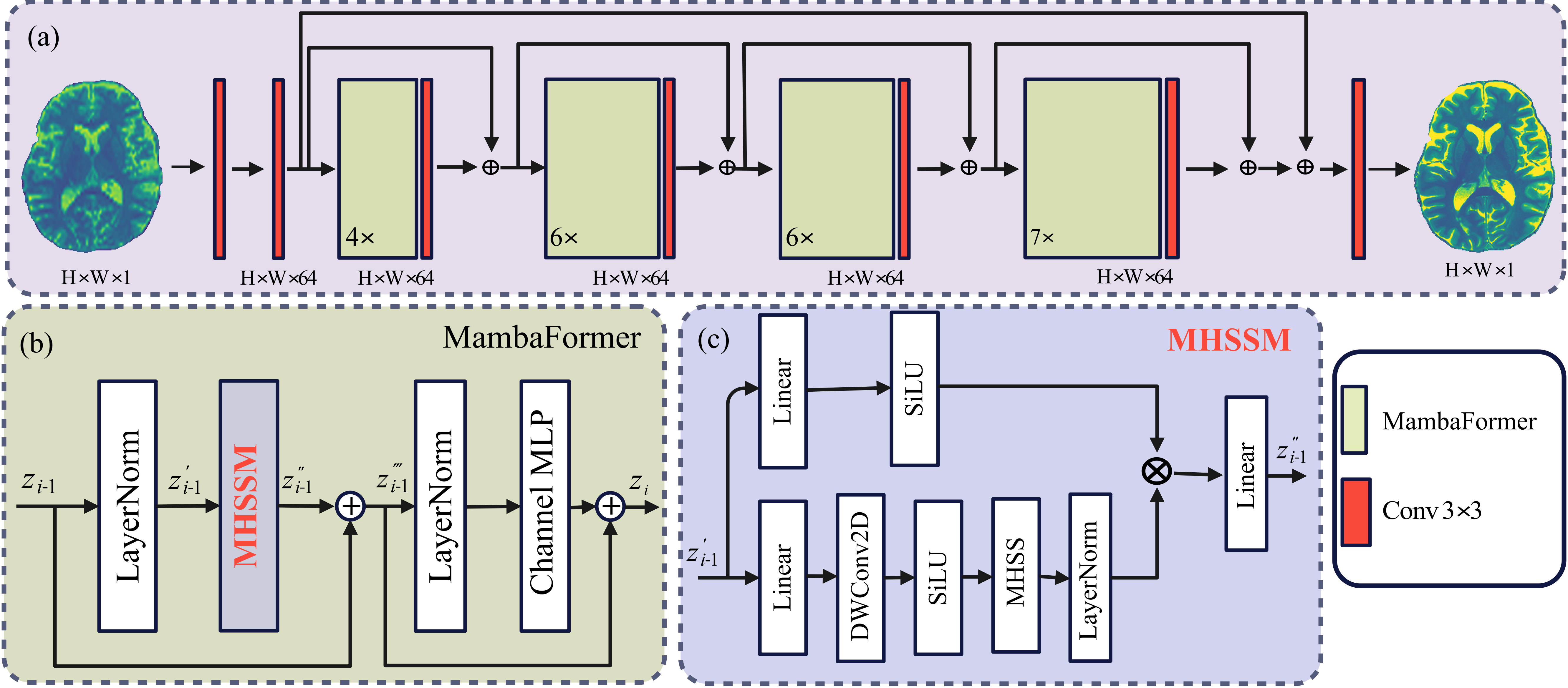}
	\caption{Proposed Vision Mamba framework for MRI super-resolution. (a) Hybrid selective scanning extracts patch sequences. (b) Each MambaFormer block integrates LN, MHSSM, and Channel MLP with residual connections. (c) Multi-head selective state-space modeling runs $m$ parallel scans with adaptive step sizes.}
	\label{fig:framework}
\end{figure}

The overall framework for reconstructing a HR image $\hr$ from a LR input $\lr$ is shown in Figure~\ref{fig:framework}. The architecture consists of four MambaFormer stages, repeated 4, 6, 6, and 7 times, respectively (Figure~\ref{fig:framework}(a)). Residual connections are employed throughout to stabilize training. Each block integrates Layer Normalization (LN), a Multi-Head SSM Mamba (MHSSM) module, and a Channel Multi-Layer Perceptron (MLP), with residual connections applied after both sub-layers (Figure~\ref{fig:framework}(b)):

\begin{equation}
	\begin{aligned}
			&z_{i-1}^{\prime\prime\prime} = z_{i-1} + \mathrm{MHSSM}(\mathrm{LN}(z_{i-1})),\
			&z_i = z_{i-1}^{\prime\prime\prime} + \mathrm{Channel , MLP}(\mathrm{LN}(z_{i-1}^{\prime\prime\prime})),
		\end{aligned}
\end{equation}
where $z_{i-1}$ and $z_i$ are the input and output features of the $i$-th block, respectively.

The Channel MLP is designed to be lightweight while preserving expressivity, inspired by recent efficient vision architectures~\cite{tolstikhin2021mlp,chen2022simple}. Given an input $x \in \mathbb{R}^{B\times C\times H\times W}$, a $1\times1$ convolution expands the channel dimension by a factor $\alpha$ (set to $\alpha=2$). The features are then split into two halves $(x_1, x_2)$, and a multiplicative gating computes $x_1 \odot x_2$. A second $1\times1$ convolution projects the result back to the original dimension:
\begin{equation}
	\mathrm{Channel \, MLP}(x) \;=\; 
	W_{\text{out}}\Big( \big(W_{\text{in}} x\big)_{1:\tfrac{C\alpha}{2}} 
	\;\odot\; \big(W_{\text{in}} x\big)_{\tfrac{C\alpha}{2}:C\alpha} \Big),
\end{equation}
where $W_{\text{in}}, W_{\text{out}}$ denote learned pointwise (1$\times$1) convolutions. This design preserves the expressivity of a two-layer MLP, but is parameter-efficient.

The MHSSM module shown in Figure~\ref{fig:framework}(c) begins with a linear projection, followed by a depthwise $3\times3$ (DWConv2D) convolution and SiLU activation for local feature mixing. The processed features are then passed into a multi-head selective state-space (MHSS) layer, which runs $m$ parallel recurrent scans with adaptive step sizes. For head $i$, the update rule is:
\begin{equation}
	h_t^{(i)} = e^{\Delta_t^{(i)} A^{(i)}} h_{t-1}^{(i)}
	+ \Phi(\Delta_t^{(i)}, A^{(i)}) B^{(i)} u_t, \qquad
	y_t^{(i)} = C^{(i)} h_t^{(i)} + D^{(i)} u_t,
\end{equation}
where $A^{(i)} = \mathrm{diag}(-e^{a^{(i)}})$ ensures stability, and $\Phi(\Delta,A)=(e^{\Delta A}-I)A^{-1}$~\cite{gu2023mamba,smith2022simplified}. The step sizes $\Delta_t^{(i)}$ are predicted by a lightweight channel MLP with a softplus reparameterization. Outputs from all heads are merged, layer-normalized, optionally gated, and projected back to the model dimension by a linear layer.

Finally, a residual connection between the input features and output features was used to stabilize the training process before projecting the features into the image channel using a $ 3\times 3 $ convolution layer. The reconstructed high-resolution image $ \hathr $ was estimated by minimizing $\ell_1 $ and learned perceptual image patch similarity (LPIPS) loss $ \ell_p $. The $\ell_1$ loss promotes accurate intensity reconstruction and reduces pixel-wise distortion, while the perceptual term encourages preservation of high-frequency structural details and perceptual realism. 

\begin{equation}\label{eq:total_loss}
	\mathcal{L} = \lambda \parallel \hathr - \hr \parallel_1 + \ell_p \left( \hathr, \hr \right)
\end{equation}
where $ \lambda = 4 $ was used to put more weight on $ \ell_1 $ loss than the $ \ell_p $. 

\subsection{Quantitative and statistical analysis}

We compared our method against six benchmark approaches: Bicubic interpolation, Pix2pix~\cite{isola2017image}, CycleGAN~\cite{zhu2017unpaired}, SPSR~\cite{ma2020structure}, I$^2$SB, SwinIR, MambaIR, and Res-SRDiff~\cite{Safari_2025_res_SRDiff}. All methods, including the proposed approach and all baselines, were trained and evaluated in a 2D slice-wise manner to ensure fair comparison using their original hyperparameters.

For our proposed method, two separate models were trained independently for the brain and prostate datasets due to their distinct anatomical characteristics, contrasts, and acquisition properties. Training was performed using the Adam optimizer with an initial learning rate of $2 \times 10^{-4}$, batch size of 8, and 30 epochs. Training was conducted on an NVIDIA A100 GPU (80GB memory) using PyTorch 2.1, requiring approximately 32 hours for the brain dataset and 18 hours for the prostate dataset.

The loss weighting parameter $\lambda = 4$ in \eqref{eq:total_loss} was selected to prioritize $\ell_1$ fidelity while incorporating perceptual guidance from the LPIPS loss. This weighting follows common practice in image restoration where perceptual losses serve as auxiliary objectives with lower relative weighting~\cite{8578166, 10.1007/978-3-319-46475-6_43}.

All baseline models were trained using their original hyperparameters as specified in their respective publications. To ensure fair comparison, all methods were trained on the same training/test splits, and for sufficient iterations to guarantee convergence.

The quality of the reconstructed HR image $\hathr$ was evaluated using four quantitative metrics: peak signal-to-noise ratio (PSNR), structural similarity index (SSIM)~\cite{wang2004image}, gradient magnitude similarity deviation (GMSD)~\cite{xue2013gradient}, and learned perceptual image patch similarity (LPIPS)~\cite{sheikh2006image}. PSNR measures residual error between the reconstructed and ground-truth images; its logarithmic scale better reflects human visual perception~\cite{Safari2023_medfusiongan}, with higher values indicating better reconstruction. SSIM, GMSD, and LPIPS assess structural similarity and perceptual quality: higher SSIM (range –1 to 1, typically close to 1 in practice) indicates better agreement, while lower GMSD and LPIPS correspond to improved perceptual fidelity.

For each metric–method combination, descriptive statistics (mean and standard deviation) were reported. To assess overall group differences, we employed the non-parametric Kruskal-Wallis omnibus test separately for each metric. When omnibus significance was detected, post-hoc pairwise comparisons were conducted using Dunn’s test with Holm correction for multiple comparisons. All statistical analyses were performed in \texttt{R} (version 4.3), with significance defined as $p < 0.05$.

These metrics were selected to capture both fidelity (PSNR, SSIM) and perceptual quality (GMSD, LPIPS), thereby providing a comprehensive evaluation of reconstruction performance. Non-parametric tests were employed because distributional assumptions of normality were not satisfied across test samples (Shapiro-Wilk test, $p<0.05$), ensuring robustness of the statistical comparisons.

\subsection{ Subjective image-quality evaluation}
\label{subsec:likert_pairwise}

A subjective image-quality evaluation was conducted on the Brain and Prostate datasets. For each dataset, 25 cases were randomly selected, and reconstructions from all compared methods were presented for evaluation. Three American Board of Radiology certified medical physicists independently assessed overall image quality using a five-point Likert scale, where scores ranged from 1 (poor image quality with substantial blur and/or artifacts) to 5 (excellent image quality with sharp anatomical detail and minimal artifacts).

Let $s_{i,m}\in\{1,\dots,5\}$ denote the Likert score assigned to method $m$ for scoring instance $i$, where each instance corresponds to a specific dataset-reader-case combination. For each dataset, method-wise descriptive statistics were computed, including mean $\pm$ standard deviation as well as median and interquartile range, as summarized in Table~\ref{tab:subjective_combined}. Although Likert scores are ordinal in nature, these descriptive summaries provide an interpretable and widely adopted approach for comparative assessment of perceived image quality.

To complement absolute scoring, pairwise preferences were derived directly from the Likert ratings. For each scoring instance $i$ and each unordered pair of methods $(a,b)$, the preferred method was determined as
\begin{equation}
	\text{winner}(i;a,b)=
	\begin{cases}
		a, & s_{i,a} > s_{i,b},\\
		b, & s_{i,b} > s_{i,a},\\
		\text{Tie}, & s_{i,a} = s_{i,b}.
	\end{cases}
\end{equation}
Aggregating across all scoring instances yielded the number of wins $W_a$, wins $W_b$, and ties $T$ for each method pair $(a,b)$. Excluding ties, the preference rate of method $a$ over method $b$ was defined as

\begin{equation}
	\pi_{a>b} = \frac{W_a}{W_a + W_b}.
\end{equation}

Statistical significance was evaluated using a two-sided exact binomial test under the null hypothesis $\pi_{a>b}=0.5$, with $n=W_a+W_b$ trials. In addition to pairwise comparisons, an overall preference rate was computed for each method by aggregating its wins and losses across all head-to-head comparisons, defined as $\text{wins}/(\text{wins}+\text{losses})$. These preference statistics are also summarized in Table~\ref{tab:subjective_combined}.

\subsection{Patient data acquisition and preprocessing}

Two datasets were employed for training and evaluation: (i) an institutional ultra-high-field 7T brain T1 MP2RAGE dataset~\cite{middlebrooks20247}, and (ii) the publicly available ProstateX axial T2w prostate cancer dataset~\cite{armato2018prostatex}. These datasets represent distinct anatomical regions and contrasts, enabling comprehensive evaluation of the proposed framework across neuroimaging and oncologic applications.

The institutional brain cohort consisted of 142 patients with confirmed multiple sclerosis. Data were retrospectively collected under Mayo Clinic IRB approval, with anonymization performed in accordance with institutional policies. The dataset was split into non-overlapping subsets for training (121 patients, 14,566 axial slices) and testing (21 patients, 2,552 slices). Imaging was performed on a 7T Siemens MAGNETOM Terra system equipped with an 8-channel transmit/32-channel receive head coil. Acquisition parameters were: TR = 4.5 s, TE = 2.2 ms, TI$_1$/TI$_2$ = 0.95/2.5 s, FA$_1$/FA$_2$ = 6$^\circ$/4$^\circ$, field-of-view (FOV) = 230 $\times$ 230 mm\textsuperscript{2}, matrix size = 288 $\times$ 288, and isotropic resolution of $0.8 \times 0.8 \times 0.8$ mm\textsuperscript{3}, with a total scan time of 8:44 minutes. Brain masks were generated from inversion-1 images using FSL BET~\cite{smith2002fast} and applied to remove extracranial signal from the T1 MP2RAGE maps. For model input, T1 maps were down-sampled by a factor of 4 in each spatial dimension, yielding a voxel size of $3.2 \times 3.2 \times 3.2$ mm\textsuperscript{3}.

From the ProstateX dataset, 334 patients were randomly selected. These were partitioned into training (268 patients, 10,480 slices) and evaluation (66 patients, 2,668 slices) sets. Data were acquired on a 1.5T Siemens scanner with imaging parameters: TR = 2.2 s, TE = 202 ms, FA = 110$^\circ$, matrix size = $256 \times 256$, and voxel size of $0.66 \times 0.66 \times 1.5$ mm\textsuperscript{3}. T2w images were down-sampled by factors of 9 in-plane and 2 through-plane, resulting in an effective voxel size of $2 \times 2 \times 3$ mm\textsuperscript{3}.

Down-sampling of both the 7T brain T1 maps and the prostate T2w images was performed in the image domain using the \texttt{SimpleITK.Resample} package (version 2.1.1)~\cite{yaniv2018simpleitk}, with linear interpolation.

All images were intensity normalized to the range $ [0, 1] $ using the 1st and 99th percentiles to minimize the influence of outliers while preserving tissue contrast. No data augmentation was applied during training to establish a controlled baseline comparison across methods without augmentation-induced variability. For slice selection, background slices with minimal anatomical content were excluded: for brain data, the first and last five slices of each volume were removed along with any slices containing $ >95\% $ background. 

\section{Results}

\subsection{Brain T1 maps}

\begin{figure}
	\centering
	\includegraphics[width=\textwidth, draft=false]{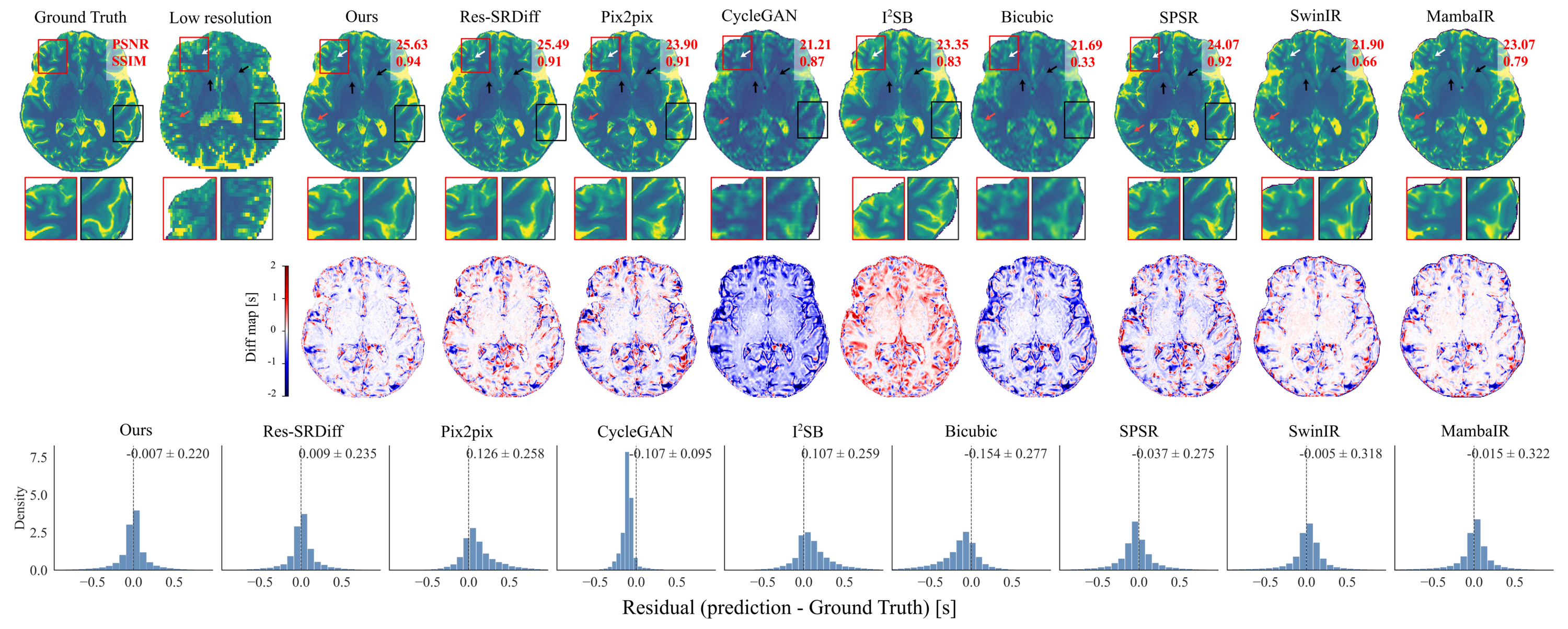}
	\caption{Qualitative comparison of super-resolution methods on 7T brain T1 MP2RAGE maps. The first row shows reconstructed images, and the second row presents magnified regions of interest. The third row displays voxel-wise signed difference maps relative to the ground-truth HR image, highlighting spatial error patterns. The bottom row shows voxel-wise residual histograms aggregated across all test cases for each method, with mean $\pm$ standard deviation reported above each distribution. White arrows indicate cortical ribbon delineation, black arrows highlight subcortical structures (caudate nucleus and putamen), and red arrows mark subtle tissue boundaries prone to reconstruction errors. Distributions centered near zero with narrower spread indicate reduced bias and improved consistency. Compared with competing approaches, the proposed method demonstrates improved boundary delineation, reduced structured residuals, and a more symmetric error distribution.}
	\label{fig:brain_qualitative}
\end{figure}

Figure~\ref{fig:brain_qualitative} shows qualitative comparisons of SR results on 7T brain T1 MP2RAGE maps. The first row displays reconstructed images, while the second row depicts difference maps relative to the ground-truth HR image. The proposed method yields the most faithful reconstruction of fine anatomical details. Subtle cortical and subcortical structures (white arrows) are more sharply delineated, and the head of the caudate nucleus and putamen (black arrows) are more accurately recovered compared to competing approaches. Both our model and Res-SRDiff demonstrate improved recovery of these regions, whereas Bicubic and CycleGAN produce pronounced blurring and loss of structural definition. Pix2pix and SPSR achieve moderate improvements but still fail to preserve fine tissue boundaries. SwinIR shows reasonable global consistency but exhibits residual blurring in cortical ribbon regions, while MambaIR captures local edge details but introduces minor inconsistencies in subcortical structures. The difference maps further support these observations: our method yields the lowest overall error relative to the ground truth, followed closely by Res-SRDiff. In contrast, Bicubic, CycleGAN, SwinIR, and MambaIR exhibit higher residual variations, reflecting suboptimal structural fidelity. Collectively, these qualitative results highlight the capability of our approach to preserve anatomically relevant features while minimizing reconstruction artifacts.

\begin{figure}[h!]
	\centering
	\includegraphics[width=\textwidth, draft=false]{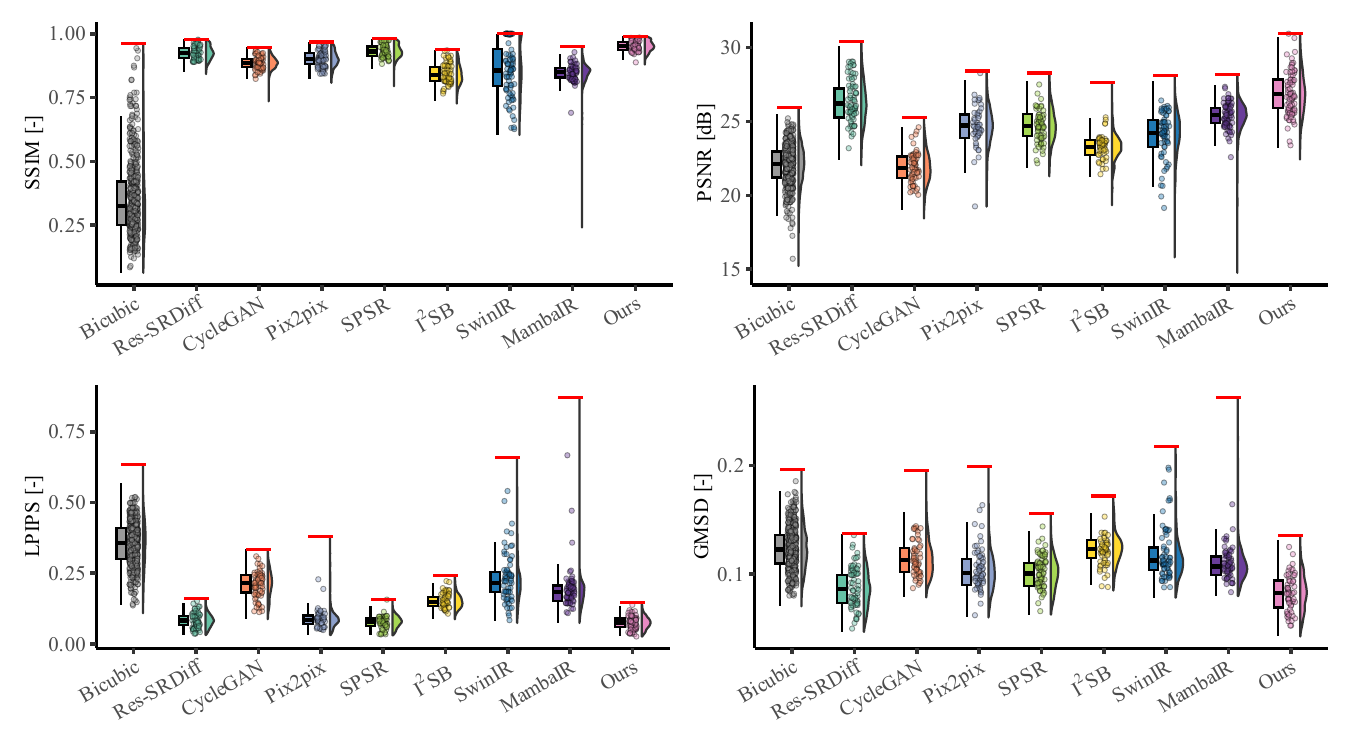}
	\caption{Quantitative comparison of super-resolution methods on the 7T brain T1 MP2RAGE dataset. For each metric (SSIM, PSNR, LPIPS, and GMSD), boxplots summarize the distribution across the test cohort (median and interquartile range), half-violins indicate the corresponding density, and points show samples. Higher SSIM/PSNR and lower LPIPS/GMSD indicate better performance.}
	\label{fig:brain_quantitative_brain}
\end{figure}

Quantitative evaluation across four image quality metrics (SSIM, PSNR, LPIPS, and GMSD) further demonstrates the superiority of the proposed method (Table~\ref{tab:qmetrics_transposed}, Figure~\ref{fig:brain_quantitative_brain}). Our model achieved the highest mean SSIM score ($0.951 \pm 0.021$), surpassing all competing techniques, with SPSR ranking second ($0.932 \pm 0.025$). For PSNR, our model reached $26.900 \pm 1.410$ dB, outperforming Res-SRDiff ($26.282 \pm 1.418$ dB) and substantially exceeding Pix2pix, SPSR ($\sim$24.7 dB), SwinIR ($24.022 \pm 1.615$ dB), and MambaIR ($25.309 \pm 1.086$ dB). In terms of perceptual similarity (LPIPS), our method obtained the lowest score ($0.076 \pm 0.022$), indicating superior perceptual fidelity compared with SPSR ($0.078 \pm 0.021$), Res-SRDiff ($0.083 \pm 0.024$), SwinIR ($0.229 \pm 0.079$), and MambaIR ($0.188 \pm 0.071$). Finally, for distortion sensitivity (GMSD), our approach again delivered the best value ($0.083 \pm 0.017$), outperforming Res-SRDiff ($0.086 \pm 0.017$), SwinIR ($0.117 \pm 0.022$), MambaIR ($0.110 \pm 0.019$), and all other baselines.

Statistical testing confirmed the robustness of these findings. The Kruskal-Wallis omnibus test revealed significant group differences for all four metrics (SSIM: $p < 0.001$; PSNR: $p < 0.001$; LPIPS: $p < 0.001$; GMSD: $p < 0.001$). Post-hoc Dunn tests with Holm-adjusted $p$-values showed that our method significantly outperformed nearly all baselines across all metrics (adjusted $p < 0.001$ in most pairwise comparisons). These results confirm that the proposed framework consistently achieves statistically significant improvements in both fidelity- and perceptual-based metrics, establishing its robustness compared to the baseline models.

To further assess perceptual image quality, we conducted a subjective evaluation using 5-point Likert scoring (Section~\ref{subsec:likert_pairwise}). The distribution of Likert scores and derived pairwise preferences for the Brain dataset are shown in Figure~\ref{fig:likert_pref}(a,c), and summarized in Table~\ref{tab:subjective_combined}.

Consistent with the quantitative findings, our method achieved the highest average Likert score ($4.27 \pm 0.70$), followed by Res-SRDiff ($3.96 \pm 0.70$). In pairwise comparisons, our approach obtained the highest overall preference rate (0.981), substantially exceeding Res-SRDiff (0.916) and all other baselines. Against Res-SRDiff, our method was preferred in 28 versus 5 non-tied comparisons ($\pi=0.848$, $p=6.6\times10^{-5}$), indicating statistically significant superiority in perceived image quality. 

These subjective findings corroborate both the qualitative observations and objective metric improvements, confirming that the proposed method provides superior anatomical fidelity and perceptual realism.

Importantly, these gains were achieved with substantially improved computational efficiency. Our model requires only 0.9M parameters and 57 GFLOPs, compared with Res-SRDiff (394M parameters, 2316 GFLOPs), SPSR (96M parameters, 871 GFLOPs), SwinIR (2M parameters, 369 GFLOPs), and MambaIR (1M parameters, 113 GFLOPs). This exceptional reduction highlights the favorable trade-off between accuracy and computational cost, making the method highly suitable for scalable clinical integration. Results are reported with two decimal places in tables for readability and three decimals in the text to emphasize fine-grained differences.

\subsection{Pelvic T2w images}

\begin{figure}[b!]
	\centering
	\includegraphics[width=\textwidth, draft=false]{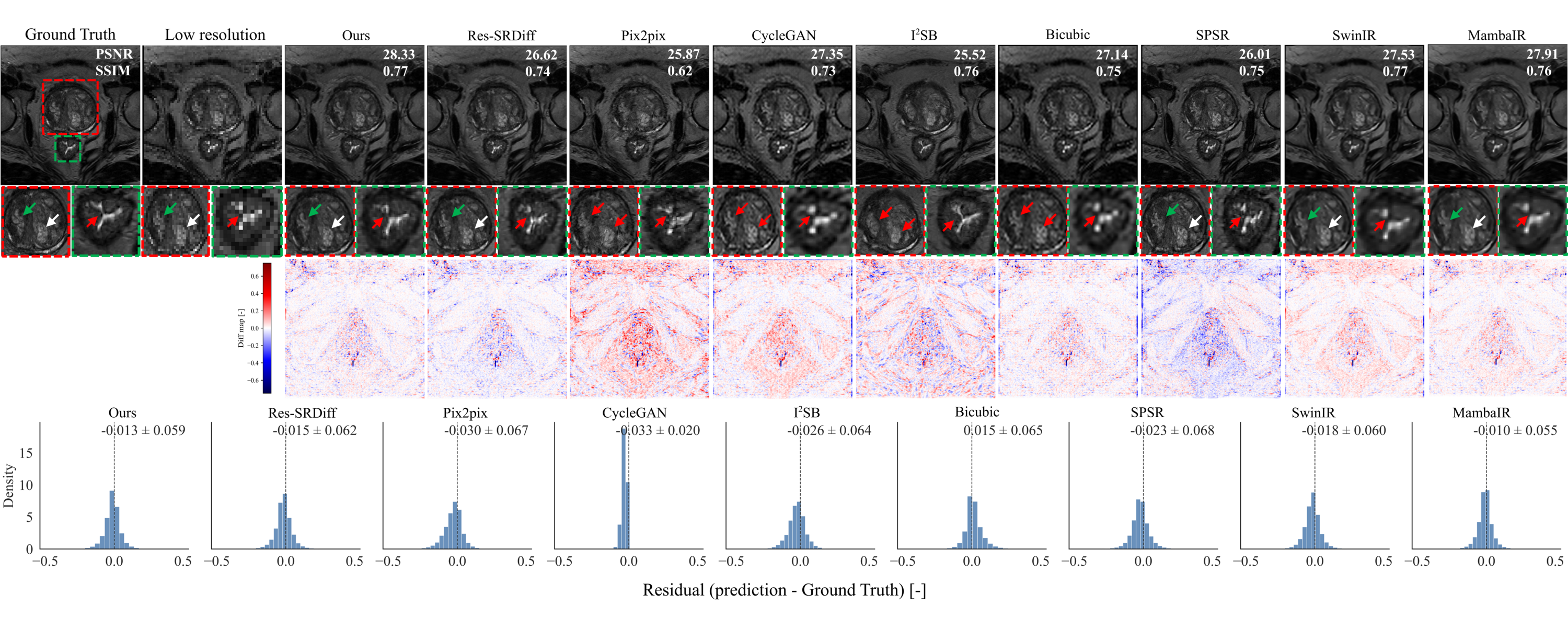}
	\caption{Visual comparison of super-resolution approaches on an axial prostate T2-weighted MRI slice. The top row displays the ground-truth image and reconstructed outputs from each method. The second row presents magnified regions of interest highlighting the prostate capsule and lesion-adjacent structures. The third row shows voxel-wise signed difference maps relative to the ground truth, illustrating spatial error patterns. The bottom row depicts voxel-wise residual histograms aggregated across the entire test cohort for each method, with mean $\pm$ standard deviation reported above each distribution. Green arrows indicate prostate capsule delineation, white arrows highlight the lesion region, and red arrows mark structures sensitive to artifact amplification or over-smoothing. Distributions centered near zero with reduced spread indicate lower bias and improved consistency. Compared with baseline approaches, the proposed method better preserves structural continuity while exhibiting a more symmetric and compact residual distribution.
	}
	\label{fig:prostate_qualitative}
\end{figure}

Figure~\ref{fig:prostate_qualitative} presents a qualitative comparison of SR results on axial T2w pelvic MRI. The ground-truth HR image (leftmost) clearly delineates anatomical boundaries, including the prostate capsule (green arrow), lesion region (white arrow), and surrounding structures (red arrow). Bicubic interpolation fails to recover fine anatomical detail, producing blurred boundaries and oversmoothed textures. GAN-based approaches such as CycleGAN and Pix2pix partially restore high-frequency components but introduce hallucinated structures and amplified noise, as seen in irregular residual patterns. SPSR demonstrates improved texture recovery, although boundary sharpness remains suboptimal. Transformer-based SwinIR shows reasonable global consistency but exhibits edge oversmoothing in fine tissue boundaries, while Mamba-based MambaIR captures local details but occasionally introduces minor artifacts in homogeneous regions. Diffusion-based methods (I\textsuperscript{2}SB and Res-SRDiff) achieve higher fidelity, yet residual artifacts and structural inconsistencies are still visible in lesion-adjacent regions.

In contrast, the proposed method reconstructs sharper anatomical edges and preserves tissue continuity, closely matching the ground-truth HR reference. Lesion boundaries and the prostate capsule are more accurately delineated, and residual maps confirm lower reconstruction error relative to competing approaches. These qualitative findings align with the quantitative results summarized in Table~\ref{tab:qmetrics_transposed}, demonstrating that our model not only enhances perceptual similarity but also reduces distortion and preserves diagnostically relevant structures.

\begin{figure}[b!]
	\centering
	\includegraphics[width=\textwidth, draft=false]{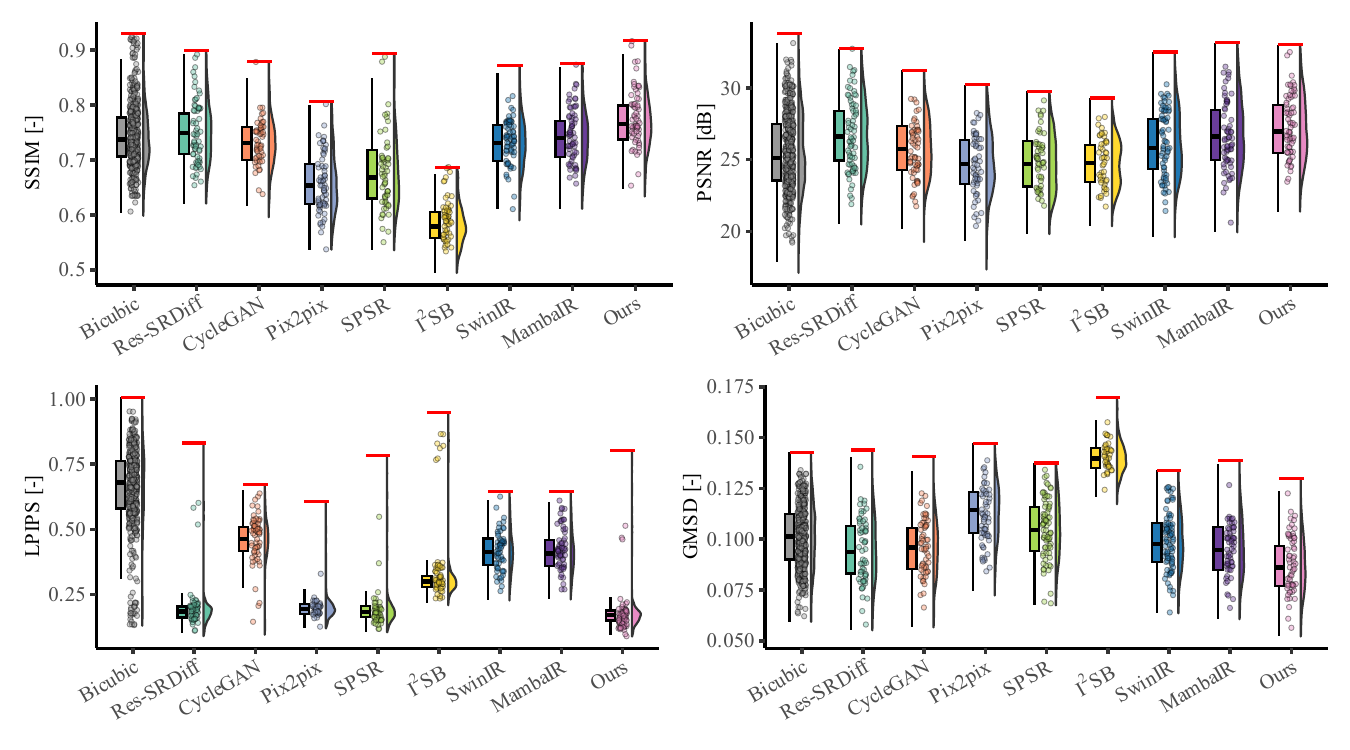}
	\caption{Quantitative comparison of super-resolution methods on the prostate T2w MRI dataset. For each metric (SSIM, PSNR, LPIPS, and GMSD), boxplots summarize the distribution across the test set (median and interquartile range), half-violins show the corresponding density, and points indicate samples. Higher SSIM/PSNR and lower LPIPS/GMSD indicate better performance.}
	\label{fig:prostate_quantitative_prostate}
\end{figure}


Quantitative comparisons on the prostate dataset are summarized in Table~\ref{tab:qmetrics_transposed} and illustrated in Figure~\ref{fig:prostate_quantitative_prostate}. Across all four evaluation metrics, our model consistently outperformed existing approaches. Specifically, it achieved the highest SSIM ($0.770 \pm 0.049$) and PSNR ($27.15 \pm 2.19$ dB), while also obtaining the lowest LPIPS ($0.184 \pm 0.096$) and GMSD ($0.083 \pm 0.013$). These results confirm that the proposed method improves both structural fidelity and perceptual quality compared with the baselines.

Figure~\ref{fig:prostate_quantitative_prostate} further illustrates these findings: Bicubic interpolation exhibits the lowest SSIM and PSNR, while GAN-based models (CycleGAN, Pix2pix, SPSR) show moderate improvements but remain inferior in perceptual similarity (higher LPIPS). Transformer-based SwinIR and Mamba-based MambaIR demonstrate intermediate performance, with SwinIR achieving better structural metrics but higher perceptual errors, and MambaIR showing competitive GMSD but suboptimal SSIM. Diffusion-based methods (Res-SRDiff, I$^2$SB) achieve higher PSNR but still underperform compared to our model in perceptual quality and distortion sensitivity. Our approach consistently achieved the best performance across all metrics, with clear margins over both GAN-, transformer-, Mamba-, and diffusion-based baselines.

To confirm statistical significance, we conducted Kruskal-Wallis omnibus tests, which revealed significant overall differences across all metrics ($p < 0.001$). Post-hoc Dunn tests with Holm correction verified that our method significantly outperformed the comparative models across all four metrics ($p < 0.001$), with no statistically significant difference compared to Res-SRDiff on LPIPS. Together, these findings demonstrate that our method provides both statistically and practically significant improvements for prostate MRI SR.

\begin{table*}[t!]
	\centering
	\caption{Quantitative results of super-resolution experiments on pelvic T2w MRI and 7T brain T1 MP2RAGE datasets, presented with metrics as rows and methods as columns. Values are mean$\pm$SD. The best scores for each metric are highlighted in bold, while the second-best are underlined. Arrows indicate the preferred direction.}
	\label{tab:qmetrics_transposed}
	\resizebox{\textwidth}{!}{
		\begin{tabular}{llccccccccc}
			\toprule
			Dataset & Metric & Bicubic & CycleGAN & Pix2pix & SPSR        &SwinIR&    MambaIR               & I\textsuperscript{2}SB & Res\mbox{-}SRDiff & Ours \\
			\midrule
			\multirow{4}{*}{Pelvic T2w MRI}
			& PSNR [dB] $\uparrow$ 
			& \mc{25.47}{2.61} & \mc{25.84}{1.96} & \mc{24.83}{2.09} & \mc{24.74}{1.96} 	& \mc{26.06}{2.25}	&	\mc{26.79}{2.35}	 & \mc{26.78}{2.35} & $ \mathbf{27.72_{\pm 2.26}} $ & \second{\mc{27.14}{2.19}} \\
			& SSIM [-] $\uparrow$
			& \second{\mc{0.75}{0.06}} & \mc{0.73}{0.05} & \mc{0.66}{0.05} & \mc{0.68}{0.07}  & \mc{0.73}{0.05}	&	\mc{0.74}{0.05}           & \mc{0.70}{0.04} & \second{\mc{0.75}{0.05}} & $ \mathbf{0.77_{\pm 0.05}} $ \\
			& GMSD [-] $\downarrow$ 
			& \second{\mc{0.10}{0.02}} & \second{\mc{0.10}{0.01}} & \mc{0.11}{0.01} & \mc{0.11}{0.01} &  \mc{0.10}{0.01}  &     \mc{0.09}{0.01}           & \mc{0.14}{0.01} & $ \mathbf{0.08_{\pm 0.02}} $ & $ \mathbf{0.08_{\pm 0.01}} $ \\
			& LPIPS [-] $\downarrow$
			& \mc{0.69}{0.15} & \mc{0.45}{0.10} & \second{\mc{0.20}{0.05}} & \second{\mc{0.20}{0.09}} &  \mc{0.42}{0.08}  &     \mc{0.41}{0.08}           & \mc{0.33}{0.13} & \mc{0.21}{0.11} & $ \mathbf{0.18_{\pm 0.10}} $\\[0.5ex]
			\hdashline
			\multirow{4}{*}{7T brain T1 MP2RAGE}
			& PSNR [dB] $\uparrow$
			& \mc{22.00}{1.37} & \mc{21.89}{1.09} & \mc{24.63}{1.32} & \second{\mc{24.76}{1.12}} &  \mc{24.02}{1.62}  &     \mc{25.31}{1.09}           & \mc{23.22}{0.98} & \second{\mc{26.28}{1.41}} & $ \mathbf{26.90_{\pm 1.41}} $ \\
			& SSIM [-] $\uparrow$
			& \mc{0.31}{0.16} & \mc{0.86}{0.02} & \mc{0.90}{0.03} & \second{\mc{0.93}{0.02}} &  \mc{0.86}{0.10}  &     \mc{0.85}{0.04}           & \mc{0.84}{0.04} & \mc{0.92}{0.03} & $ \mathbf{0.95_{\pm 0.02}} $ \\
			& GMSD [-] $\downarrow$
			& \mc{0.12}{0.02} & \mc{0.12}{0.02} & \second{\mc{0.10}{0.02}} & \second{\mc{0.10}{0.01}} &  \mc{0.12}{0.02}  &     \mc{0.11}{0.02}           & \mc{0.12}{0.01} & \best{\mc{0.07}{0.02}} & $ \mathbf{0.07_{\pm 0.02}} $ \\
			& LPIPS [-] $\downarrow$
			& \mc{0.38}{0.07} & \mc{0.21}{0.05} & \second{\mc{0.09}{0.04}} & \best{\mc{0.08}{0.02}} &  \mc{0.23}{0.08}  &     \mc{0.19}{0.07}           & \mc{0.15}{0.03} & $ \mathbf{0.08_{\pm 0.02}} $ & $ \mathbf{0.08_{\pm 0.02}} $ \\[0.5ex]
			\hdashline
			\multirow{2}{*}{Model size} 
			& Flops [G] & x &  18 &  9 & 871 &  369  &     113           & 497 & 1316 & 57 \\
			& Params [m] & x &  89 &  57 & 96 &  2  &     1           & 114 & 394 & 0.9 \\
			\bottomrule
		\end{tabular}
	}
\end{table*}

\begin{figure}[t]
	\centering
	\includegraphics[width=0.9\linewidth]{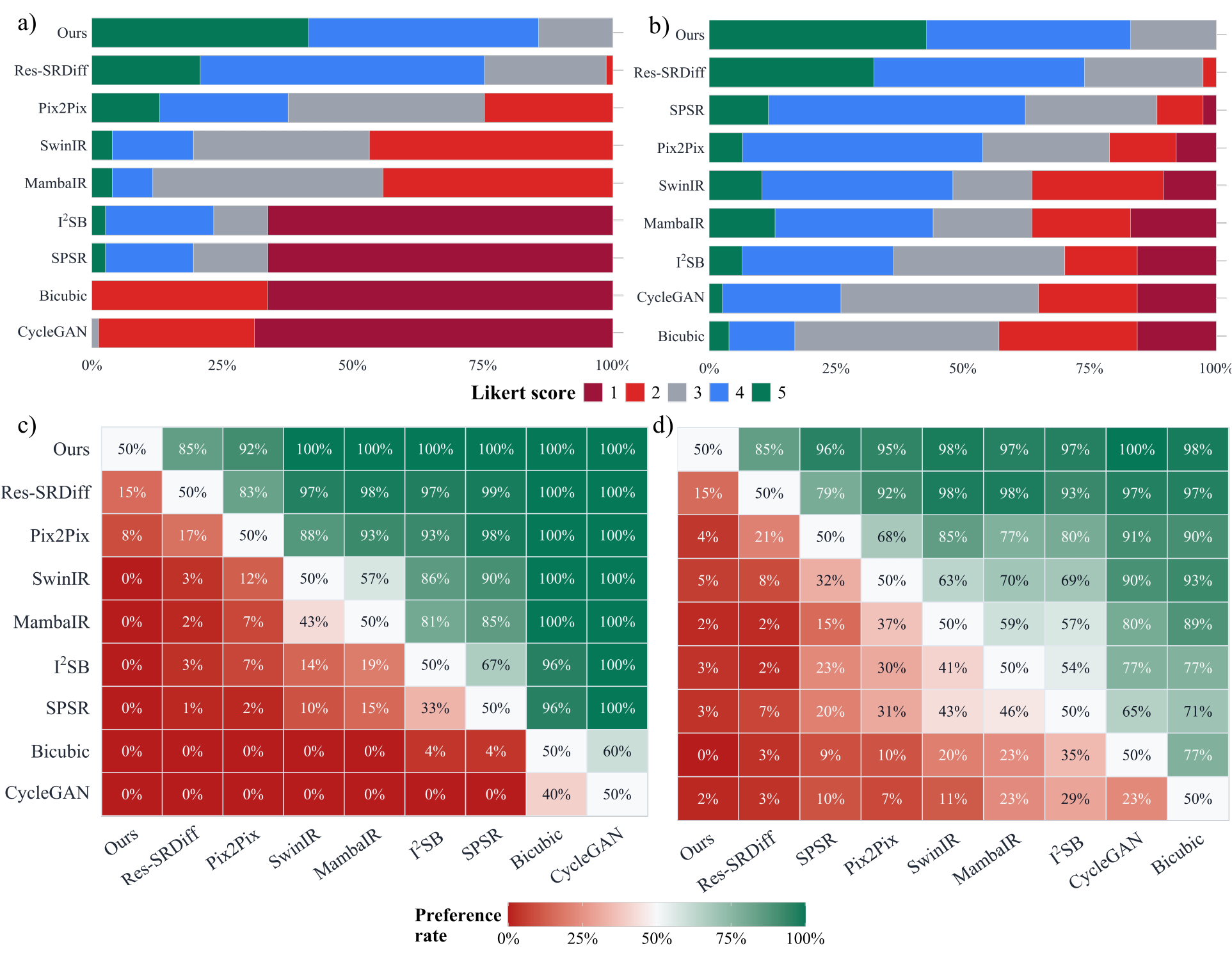}
	\caption{{Subjective Likert image-quality scoring and derived pairwise preferences.}
		Panels (a,c) correspond to Brain and panels (b,d) correspond to Prostate. (a-b) Stacked distributions of Likert scores (1-5) per method (higher is better). (c-d) Pairwise preference-rate matrices derived from Likert scores; each cell reports the preference rate (row method preferred over column method), computed as wins/(wins+losses) excluding ties; the diagonal is 50\%.}
	\label{fig:likert_pref}
\end{figure}

Subjective evaluation on the Prostate dataset further supports these results. As illustrated in Figure~\ref{fig:likert_pref}(b,d) and summarized in Table~\ref{tab:subjective_combined}, our method achieved the highest Likert score ($4.26 \pm 0.73$), outperforming Res-SRDiff ($4.03 \pm 0.81$) and all other competing methods.

Pairwise preference analysis demonstrated a consistent advantage of the proposed approach, with the highest overall preference rate (0.966), compared to 0.888 for Res-SRDiff. Direct comparison between the two strongest methods showed that our model was preferred in 22 versus 4 non-tied evaluations ($\pi=0.846$, $p=5.3\times10^{-4}$). 

These subjective outcomes align with both quantitative metrics and qualitative visual assessment, indicating that the proposed framework improves structural delineation and perceptual fidelity in clinically relevant regions.

\begin{table*}[t]
	\centering
	\caption{Subjective image-quality evaluation using 5-point Likert scoring and derived pairwise preference rates.
		Likert values are mean$\pm$SD (higher is better).
		Preference values are wins/(wins+losses) aggregated across all head-to-head comparisons (ties excluded).
		The best value per row is highlighted in bold and the second-best is underlined.}
	\label{tab:subjective_combined}
	\resizebox{\textwidth}{!}{
		\begin{tabular}{llccccccccc}
			\toprule
			Dataset & Metric 
			& Bicubic & CycleGAN & Pix2Pix & SPSR & SwinIR & MambaIR & I\textsuperscript{2}SB & Res-SRDiff & Ours \\
			\midrule
			
			\multirow{2}{*}{Brain}
			& Likert (1--5) $\uparrow$
			& 1.34$\pm$0.48
			& 1.32$\pm$0.50
			& 3.27$\pm$0.98
			& 1.89$\pm$1.31
			& 2.77$\pm$0.85
			& 2.71$\pm$0.78
			& 1.93$\pm$1.36
			& \underline{3.96$\pm$0.70}
			& \textbf{4.27$\pm$0.70} \\
			
			& Preference $\uparrow$
			& 0.011
			& 0.005
			& 0.766
			& 0.182
			& 0.591
			& 0.563
			& 0.216
			& \underline{0.916}
			& \textbf{0.981} \\[0.5ex]
			
			\hdashline
			
			\multirow{2}{*}{Prostate}
			& Likert (1--5) $\uparrow$
			& 2.62$\pm$1.03
			& 2.79$\pm$1.06
			& 3.31$\pm$1.05
			& \underline{3.59$\pm$0.91}
			& 3.12$\pm$1.22
			& 3.04$\pm$1.31
			& 2.98$\pm$1.16
			& 4.03$\pm$0.81
			& \textbf{4.26$\pm$0.73} \\
			
			& Preference $\uparrow$
			& 0.116
			& 0.161
			& 0.534
			& \underline{0.660}
			& 0.399
			& 0.345
			& 0.343
			& 0.888
			& \textbf{0.966} \\[0.5ex]
			
			\hdashline
			
			\multirow{2}{*}{Combined}
			& Likert (1--5) $\uparrow$
			& 1.98$\pm$1.03
			& 2.06$\pm$1.11
			& 3.29$\pm$1.01
			& 2.74$\pm$1.41
			& 2.94$\pm$1.06
			& 2.88$\pm$1.09
			& 2.45$\pm$1.36
			& \underline{4.00$\pm$0.76}
			& \textbf{4.26$\pm$0.71} \\
			
			& Preference $\uparrow$
			& 0.062
			& 0.077
			& 0.664
			& 0.422
			& 0.511
			& 0.470
			& 0.281
			& \underline{0.904}
			& \textbf{0.974} \\
			
			\bottomrule
		\end{tabular}
	}
\end{table*}

Importantly, these performance gains were achieved with substantially improved computational efficiency. As shown in Table~\ref{tab:qmetrics_transposed}, our method requires only 57 GFLOPs and 0.9M parameters, compared to 1316 GFLOPs and 394M parameters for Res-SRDiff and 871 GFLOPs with 96M parameters for SPSR. This demonstrates that the proposed model achieves state-of-the-art image quality at a fraction of the computational and memory cost, advantages that render it well suited for future clinical validation and integration.

\subsection{Ablation Study} \label{subsec:ablation}

To empirically validate the architectural and training design choices of the proposed framework, we performed a controlled ablation study isolating the effects of block repetition depth, scan strategy and multiplicity, feed-forward expansion factor, and loss composition. All experiments were conducted under identical training and evaluation protocols, with only a single component modified relative to the baseline configuration. The baseline model employs a four-stage topology with block repetition configuration $[4,6,6,7]$, feed-forward expansion factor $\alpha=2.0$, multi-path diagonal scanning (scan\_type = diagonal, scan\_count = 8), and a composite loss defined as $\mathcal{L} = 4\mathcal{L}_1 + \mathcal{L}_{\text{LPIPS}}$. No adversarial term is used in the optimization objective.

To evaluate the influence of representational depth within each stage, we first modified the block repetition configuration while keeping the four-stage topology unchanged. A lighter configuration ($[3,4,4,5]$, denoted {blocks\_light}) reduced PSNR from $27.14 \pm 2.19$ dB to $26.46 \pm 2.17$ dB and SSIM from $0.77 \pm 0.05$ to $0.74 \pm 0.05$. Increasing intra-stage depth beyond the baseline ($[5,7,7,8]$, denoted {block\_heavy}) did not yield measurable improvement ($26.48 \pm 2.13$ dB, $0.74 \pm 0.05$ SSIM). These results indicate that the selected block configuration provides a balanced trade-off between representational capacity and optimization stability, and that additional depth alone does not monotonically improve reconstruction fidelity.

We next examined the contribution of scan multiplicity and traversal topology. Reducing the number of diagonal scan paths from 8 to 4 ({scan4}) resulted in a performance decrease to $26.45 \pm 2.14$ dB PSNR and $0.74 \pm 0.05$ SSIM, demonstrating that multi-path aggregation contributes to improved reconstruction accuracy. Replacing diagonal traversal with a zigzag scan pattern while retaining eight scan paths ({zigzag\_scan}) also reduced performance ($26.51 \pm 2.09$ dB, $0.74 \pm 0.05$ SSIM). Together, these findings support the use of multi-path diagonal scanning as the most effective traversal strategy among the evaluated alternatives.

To assess the role of channel expansion within the feed-forward network, we reduced the expansion factor from $\alpha=2.0$ to $\alpha=1.0$ (\textit{alpha1}). This modification led to a decrease in PSNR to $26.56 \pm 2.16$ dB and SSIM to $0.75 \pm 0.05$, suggesting that feature expansion enhances the representational richness of intermediate activations and contributes to improved reconstruction performance.

Finally, we evaluated the effect of perceptual supervision by removing the LPIPS term and optimizing solely with $\mathcal{L} = 4\mathcal{L}_1$ ({l1\_only}). This configuration yielded $26.97 \pm 2.32$ dB PSNR and $0.77 \pm 0.05$ SSIM. Although structural similarity remained comparable, the composite loss achieved the highest overall PSNR, indicating that combining pixel-wise and perceptual supervision improves fidelity beyond pure $\mathcal{L}_1$ optimization.

Across all controlled modifications, the baseline configuration consistently achieved the best quantitative performance. These results empirically substantiate the selected block repetition configuration, multi-path diagonal scanning strategy, expansion factor, and composite loss formulation.

\begin{table}[t]
	\centering
	\caption{Ablation study results. Baseline corresponds to block configuration $[4,6,6,7]$, $\alpha=2.0$, diagonal scanning with 8 paths, and $\mathcal{L}=4\mathcal{L}_1 + \mathcal{L}_{\text{LPIPS}}$. Values are mean$\pm$SD.}
	\label{tab:ablation}
	\begin{tabular}{lcc}
		\toprule
		Configuration & PSNR [dB] $\uparrow$ & SSIM $\uparrow$ \\
		\midrule
		blocks\_light & $26.46 \pm 2.17$ & $0.74 \pm 0.05$ \\
		block\_heavy  & $26.48 \pm 2.13$ & $0.74 \pm 0.05$ \\
		scan4         & $26.45 \pm 2.14$ & $0.74 \pm 0.05$ \\
		zigzag\_scan  & $26.51 \pm 2.09$ & $0.74 \pm 0.05$ \\
		alpha1        & $26.56 \pm 2.16$ & $0.75 \pm 0.05$ \\
		l1\_only      & $26.97 \pm 2.32$ & $0.77 \pm 0.05$ \\
		\midrule
		\textbf{Baseline (Ours)} & $\mathbf{27.14 \pm 2.19}$ & $\mathbf{0.77 \pm 0.05}$ \\
		\bottomrule
	\end{tabular}
\end{table}

\section{Discussion}

MRI remains one of the most versatile imaging modalities in both clinical and research applications, offering exceptional soft-tissue contrast and flexible imaging capabilities without exposing patients to ionizing radiation. However, its inherently lengthy acquisition times reduce scanner throughput and increase the likelihood of patient motion, leading to artifacts. This trade-off between spatial resolution and acquisition efficiency often necessitates compromises: increasing voxel size can shorten scan duration but reduces resolution and image sharpness, with diagnostic performance further degraded by partial volume effects~\cite{mao2023high}.

In this study, we developed an efficient Vision Mamba model to reconstruct HR images from LR inputs. By leveraging hybrid scanning directions, our approach mitigates voxel forgetting (as illustrated in Figure~\ref{fig:selective_scan}(a)-(b)), while the multi-head selective state-space module (MHSSM; Figure~\ref{fig:framework}(c)-(d)) captures both local and long-range dependencies. Despite requiring fewer than one million parameters, our study achieves competitive or superior performance compared with much larger models, including transformer-based SwinIR and Mamba-based MambaIR.

Our findings demonstrate that the model consistently delivers high-quality reconstructions across both brain and prostate MRI, while maintaining remarkable computational efficiency. In brain MRI, the method excelled at recovering fine cortical and subcortical structures, including the caudate nucleus and putamen, which are critical for neuroimaging analyses. As shown in Figure~\ref{fig:brain_qualitative}, subtle tissue boundaries were more sharply delineated compared with competing methods. Bicubic interpolation and CycleGAN exhibited substantial structural loss, while Pix2pix and SPSR partially restored anatomical features but failed to preserve finer details. Transformer-based SwinIR showed reasonable global consistency but suffered from edge blurring, and Mamba-based MambaIR captured local details but introduced inconsistencies in complex anatomical regions. Although Res-SRDiff provided competitive recovery, its perceptual fidelity remained inferior to our model.

Quantitative evaluation further supported these observations. The proposed method achieved the highest SSIM ($0.951 \pm 0.021$) and PSNR ($26.900 \pm 1.410$ dB) on the 7T brain dataset, while also recording the lowest LPIPS ($0.076 \pm 0.022$) and GMSD ($0.083 \pm 0.017$). Statistical testing confirmed significant differences across methods ($p < 0.001$ for all metrics), with Dunn post-hoc tests verifying that our model outperformed all baselines, including SwinIR and MambaIR. Importantly, these improvements were achieved with only 57 GFLOPs and 0.9M parameters, compared to Res-SRDiff (2316 GFLOPs, 394M parameters), SPSR (871 GFLOPs, 96M parameters), SwinIR (369 GFLOPs, 2M parameters), and MambaIR (113 GFLOPs, 1M parameters). These results suggest that architectural efficiency, rather than brute computational scale, can yield clinically meaningful advances.

A similar trend was observed in prostate MRI SR. As shown in Figure~\ref{fig:prostate_qualitative}, our method reconstructed sharper prostate capsule boundaries and more faithfully preserved lesion morphology compared with competing models. GAN-based approaches partially restored textures but often introduced hallucinations or amplified noise, limiting diagnostic reliability. Transformer-based SwinIR maintained global consistency but oversmoothed fine tissue boundaries, while Mamba-based MambaIR captured local edge details but occasionally produced artifacts in homogeneous regions. Diffusion-based methods achieved higher fidelity than GANs, but residual artifacts were still evident, especially near lesion-adjacent regions. In contrast, our approach minimized reconstruction error and preserved clinically relevant structures.

The quantitative results corroborated these findings: our method achieved the best SSIM ($0.770 \pm 0.049$) and PSNR ($27.15 \pm 2.19$ dB), while also recording the lowest LPIPS ($0.184 \pm 0.096$) and GMSD ($0.083 \pm 0.013$). These values significantly outperformed Bicubic, GAN-based, transformer-based (SwinIR), Mamba-based (MambaIR), and diffusion-based methods ($p < 0.001$ for all metrics), with only marginal differences relative to Res-SRDiff on LPIPS. Consistent with results of the brain dataset, the improvements were realized with substantially lower computational costs, underscoring the method's suitability for large-scale or real-time deployment.

Notably, the dispersion of PSNR and SSIM values was tighter in the brain dataset compared with the prostate cohort. This likely reflects differences in acquisition characteristics and anatomical variability. The 7T MP2RAGE brain dataset consists of isotropic high-resolution volumes acquired under a relatively uniform protocol, with consistent structural anatomy across subjects. In contrast, the 1.5T prostate dataset exhibits greater inter-patient anatomical variability, heterogeneous lesion presentation, anisotropic resolution, and more aggressive downsampling factors. These factors increase reconstruction difficulty and contribute to broader metric variability in the prostate experiments.

Taken together, the consistent superiority of our method across two anatomically and contrast-wise distinct datasets demonstrates both its robustness and generalizability. By combining high fidelity, perceptual accuracy, and lightweight design, the proposed model offers a practical step toward integrating SR into clinical MRI workflows, where accuracy must be balanced with efficiency.

An important consideration in SR is the well-known perception-distortion trade-off, whereby models optimized for perceptual sharpness may introduce hallucinated details, while distortion-minimizing models tend to produce smoother but potentially less visually sharp reconstructions. In our framework, the dominant $\ell_1$ loss weighting ($\lambda = 4$) constrains structural deviations from the ground truth, while the perceptual term encourages preservation of anatomically consistent high-frequency information. Unlike adversarial approaches, our method does not rely on discriminator-driven texture synthesis, which may reduce the risk of artificial structure amplification. Nevertheless, as highlighted in Figure~\ref{fig:brain_qualitative}, subtle alterations may still occur, and careful validation remains necessary before clinical deployment.

This study has several limitations that should be acknowledged. First, although both datasets consist of full 3D volumes, our framework uses 2D slice-based processing. This design choice, while enabling computational efficiency and alignment with established benchmarking practices, does not explicitly enforce volumetric continuity across adjacent slices. While all compared methods were evaluated under identical 2D conditions to ensure fair comparison, the independent processing of slices may introduce subtle inconsistencies that could affect downstream 3D applications such as volumetric segmentation or radiomics analysis. Future work should investigate 3D architectures or incorporate inter-slice regularization to better preserve volumetric coherence for applications requiring strict slice-to-slice continuity.

Second, although our method demonstrated significant gains in both fidelity- and perceptual-based metrics, the evaluation was limited to two datasets with specific contrasts (7T T1 MP2RAGE brain maps and 1.5T axial T2w prostate images). Both cohorts were acquired from a single vendor and restricted to specific field strengths, and therefore do not represent the diversity of scanners and acquisition protocols encountered in routine clinical practice. Additional validation across multi-vendor systems, including widely used 3T platforms, as well as across broader anatomical regions and imaging contrasts, is necessary to establish more comprehensive generalizability. In line with recent work on foundation models in medical imaging, which emphasize the importance of large-scale and heterogeneous training data, extending our framework to more diverse datasets will be an important next step~\cite{WANG2026103992,wang2025unifyingbiomedicalvisionlanguageexpertise}.

Third, our downsampling approach uses linear interpolation to generate low-resolution inputs, which represents a simplified degradation model. Clinical MRI resolution limitations often result from more complex processes including anisotropic blurring, partial volume effects, and sequence-specific noise. While this choice enables clear benchmarking against prior work that uses similar simplifications~\cite{liang2021swinir,10.1007/978-3-031-72649-1_13}, future validation should consider more realistic degradation models or physically-acquired paired low-resolution/high-resolution data to better assess clinical applicability.

Finally, despite the clear improvements in structural fidelity and perceptual similarity, we note that SR can, in rare cases, introduce subtle alterations to fine anatomical structures, as indicated by the red arrows in Figure~\ref{fig:brain_qualitative}; to further assess this behavior, worst-case examples identified by the lowest PSNR values are provided in Figure~S1 of the Supplementary Material, where deviations remain spatially localized and anatomically consistent. While our residual analyses suggest lower error rates compared to competing models, such artifacts may still carry clinical consequences. Future directions should include uncertainty quantification and structure-preserving constraints to safeguard against the inadvertent modification of clinically relevant features.

\section{Conclusion}

This study presented a state space-driven framework for efficient and effective MRI SR from low-resolution inputs. The proposed method integrates multi-head selective state space modules with a lightweight channel design, enhancing computational efficiency compared with conventional transformer-based models, including SwinIR and MambaIR. Experimental results demonstrate that the framework can generate high-quality, high-resolution MRI images within clinically practical time frames. Owing to its reduced computational cost and scalability, the proposed approach holds strong potential for translation and deployment across diverse clinical imaging environments.

%

%

\bibliographystyle{unsrt}
\bibliography{./sr_mamba_v01.bib}      

\end{document}